\documentclass[sigconf, screen, nonacm, balance=false]{acmart}

\usepackage{enumitem}
\usepackage{nicefrac}
\usepackage[ruled]{algorithm2e}
\usepackage{tabularray}

\setcopyright{none}

\let\maketitlesup\maketitle
\usepackage{xpatch}
\xpatchcmd{\maketitlesup}{\@mkteasers}{}{}{}
\xpatchcmd{\maketitlesup}{\@mkabstract}{}{}{}

\UseTblrLibrary{booktabs}

\newcommand{\AGG}{\textsc{Agg}}
\newcommand{\UPD}{\textsc{Upd}}
\newcommand{\MLP}{\textsc{MLP}}
\newcommand{\NAP}{\textsc{NAP}}
\newcommand{\etal}{\textit{et~al}.}
\newcommand{\norm}[2][2]{\lVert#2\rVert_{#1}}

\def\name{\textsc{ProGAP}}
\def\GraphSAGE{\textsc{GraphSAGE}}
\def\EdgeRand{\textsc{EdgeRand}}

\AtBeginDocument{}%
\AtBeginDocument{}%
\AtBeginDocument{}%
\AtBeginDocument{}%
\AtBeginDocument{}%
\AtBeginDocument{}%
\AtBeginDocument{}%
\AtBeginDocument{}%
\AtBeginDocument{}%

\begin{document}

\title{\name{}: Progressive Graph Neural Networks with Differential Privacy Guarantees}

\author{Sina Sajadmanesh}
\email{sina.sajadmanesh@epfl.ch}
\affiliation{%
  \institution{Idiap Research Institute and EPFL}
  \country{Switzerland}
}

\author{Daniel Gatica-Perez}
\email{daniel.gatica-perez@epfl.ch}
\affiliation{%
  \institution{Idiap Research Institute and EPFL}
  \country{Switzerland}
}
\begin{abstract}

  Graph Neural Networks (GNNs) have become a popular tool for learning on graphs, but their widespread use raises privacy concerns as graph data can contain personal or sensitive information. Differentially private GNN models have been recently proposed to preserve privacy while still allowing for effective learning over graph-structured datasets. However, achieving an ideal balance between accuracy and privacy in GNNs remains challenging due to the intrinsic structural connectivity of graphs. In this paper, we propose a new differentially private GNN called \name{} that uses a progressive training scheme to improve such accuracy-privacy trade-offs. Combined with the aggregation perturbation technique to ensure differential privacy, \name{} splits a GNN into a sequence of overlapping submodels that are trained progressively, expanding from the first submodel to the complete model. Specifically, each submodel is trained over the privately aggregated node embeddings learned and cached by the previous submodels, leading to an increased expressive power compared to previous approaches while limiting the incurred privacy costs. We formally prove that \name{} ensures edge-level and node-level privacy guarantees for both training and inference stages, and evaluate its performance on benchmark graph datasets. Experimental results demonstrate that \name{} can achieve up to 5-10\% higher accuracy than existing state-of-the-art differentially private GNNs. Our code is available at \href{https://github.com/sisaman/ProGAP}{https://github.com/sisaman/ProGAP}.


\end{abstract}

\maketitle

\section{Introduction}\label{sec:intro}

Graph Neural Networks (GNNs) have emerged as a powerful tool for learning from graph-structured data, and their popularity has surged due to their ability to achieve impressive performance in a wide range of applications, including social network analysis, drug discovery, recommendation systems, and traffic prediction~\cite{ying2018graph, gkalelis2021objectgraphs, jiang2021graph, 9378164, ahmedt2021graph}. GNNs excel at learning from the structural connectivity of graphs by iteratively updating node embeddings through information aggregation and transformation from neighboring nodes, making them well-suited for tasks such as node classification, graph classification, and link prediction~\cite{kipf2017semi, zhang2018link, xu2018how, corso2020pna, hamilton2017inductive, zhang2021labeling}. 
However, as with many data-driven approaches, GNNs can expose individuals to privacy risks when applied to graph data containing sensitive information, such as social connections, medical records, and financial transactions~\cite{qiu2018deepinf, wang2021review}. Recent studies have shown that various attacks, such as link stealing, membership inference, and node attribute inference, can successfully break the privacy of graph datasets~\cite{olatunji2021membership, he2021node, he2021stealing, wu2021linkteller}, posing a significant challenge for the practical use of GNNs in privacy-sensitive applications.

To address the privacy concerns associated with GNNs, researchers have recently studied \emph{differential privacy (DP)}, a well-established mathematical framework that provides strong privacy guarantees, usually by adding random noise to the data~\cite{dwork2006calibrating, dwork2008differential}. However, applying DP to GNNs is very challenging due to the complex structural connectivity of graphs, rendering traditional private learning methods, such as differentially private stochastic gradient descent (DP-SGD)~\cite{abadi2016deep}, infeasible~\cite{daigavane2021node,sajadmanesh2023gap, ayle2023training}. Recently, the \emph{aggregation perturbation} (AP) approach~\cite{sajadmanesh2023gap} has emerged as a state-of-the-art technique for ensuring DP in GNNs. Rather than perturbing the model gradients as done in the standard DP-SGD algorithm and its variants, this method perturbs the aggregate information obtained from the GNN neighborhood aggregation step. Consequently, such perturbations can obfuscate the presence of a single edge, which is called \emph{edge-level privacy}, or a single node and all its adjacent edges, referred to as \emph{node-level privacy}~\cite{raskhodnikova2016differentially}.

The key limitation of AP is its incompatibility with standard GNN architectures due to the high privacy costs it entails~\cite{sajadmanesh2023gap}. This is because conventional GNN models constantly query the aggregation functions with every update to the model parameters, which necessitates the re-perturbation of all aggregate outputs at every training iteration to ensure DP, leading to a significant increase in privacy costs. To mitigate this issue, Sajadmanesh~\etal~\cite{sajadmanesh2023gap} proposed a method called GAP, which decouples the aggregation steps from the model parameters. In GAP, node features are recursively aggregated first, and then a classifier is learned over the resulting perturbed aggregations, enabling DP to be maintained without incurring excessive privacy costs. Due to having non-trainable aggregations, however, such decoupling approaches reduce the representational power of the GNN~\cite{fey2021gnnautoscale}, leading to suboptimal accuracy-privacy trade-offs.

In the face of these challenges, we present a novel differentially private GNN, called \emph{``\textbf{Pro}gressive \textbf{G}NN with \textbf{A}ggregation \textbf{P}erturbation''} (\name{}). Our new method uses the same AP technique as in GAP to ensure DP. However, instead of decoupling the aggregation steps from the learnable modules, \name{} adopts a multi-stage, progressive training paradigm to surmount the formidable privacy costs associated with AP. Specifically, \name{} converts a $K$-layer GNN model into a sequence of overlapping submodels, where the $i$-th submodel comprises the first $i$ layers of the model, followed by a lightweight supervision head layer with softmax activation that utilizes node labels to guide the submodel's training. Starting with the shallowest submodel, \name{} then proceeds progressively to train deeper submodels, each of which is referred to as a training stage. At every stage, the learned node embeddings from the preceding stage are aggregated, perturbed, and then cached to save privacy budget, allowing \name{} to learn a new set of private node embeddings. Ultimately, the last stage's embeddings are used to generate final node-wise predictions.

The proposed progressive training approach overcomes the high privacy costs of AP by allowing the perturbations to be applied only once per stage rather than at every training iteration. \name{} also maintains a higher level of representational power compared to GAP, as the aggregation steps now operate on the learned embeddings from the preceding stages, which are more expressive than the raw node features. Moreover, we prove that \name{} retains all the benefits of GAP, such as edge- and node-level privacy guarantees and zero-cost privacy at inference time. We evaluate \name{} on five node classification datasets, including Facebook, Amazon, and Reddit, and demonstrate that it can achieve up to 10.4\% and 5.5\% higher accuracy compared to GAP under edge- and node-level DP with an epsilon of 1 and 8, respectively.
% \begin{itemize}[leftmargin=1em]
%   \item We propose \name{}, a novel differentially private GNN framework that combines the aggregation perturbation technique with progressive training to mitigate the high privacy costs associated with the aggregation perturbation method, with a higher expressive power for attaining better accuracy-privacy trade-offs.
%   \item We prove that \name{} provides similar privacy guarantees as GAP, including edge- and node-level privacy guarantees and zero-cost privacy at inference time.
%   \item We conduct extensive experiments on a variety of node classification datasets and demonstrate that \name{} achieves state-of-the-art performance under both edge- and node-level privacy in terms of accuracy-privacy trade-off.
% \end{itemize}

% The paper is organized as follows. \autoref{sec:related} provides an overview of related work on differentially private GNNs. \autoref{sec:background} presents the necessary background and preliminaries on GNNs and differential privacy, which are crucial for understanding the proposed method. In \autoref{sec:method}, we formally describe \name{} and analyze its privacy guarantees. \autoref{sec:setup} and \autoref{sec:results} present the experimental setup and results, respectively. Finally, \autoref{sec:conclusion} concludes the paper and discusses future directions.

\section{Related Work}\label{sec:related}

Several recent studies have investigated differential privacy (DP) to provide formal privacy guarantees in various GNN learning settings. For example, Sajadmanesh and Gatica-Perez~\cite{sajadmanesh2021locally} propose a locally private GNN for a distributed learning environment, where node features and labels remain private, while the GNN training is federated by a central server with access to graph edges. Lin~\etal~\cite{lin2022towards} also introduce a locally private GNN, called \textsc{Solitude}, that preserves edge privacy in a decentralized graph, where each node keeps its own private connections. However, both of these approaches use local differential privacy~\cite{kasiviswanathan2011can}, which operates under a different problem setting from our method. % On the global DP side, Ayle~\etal~\cite{ayle2023training} propose a private GNN learning method by training a GNN model with DP-SGD over disjoint subgraphs of the input graph extracted via random walk-based algorithms. However, their approach only guarantees feature-level DP and does not preserve the privacy of edges. 

Other approaches propose edge-level DP algorithms for GNNs. Wu \etal~\cite{wu2021linkteller} developed an edge-level private method that modifies the input graph directly through randomized response or the Laplace mechanism, followed by training a GNN on the resulting noisy graph. In contrast, Kolluri~\etal~\cite{kolluri2022lpgnet} propose \textsc{LPGNet}, which adopts a tailored neural network architecture. Instead of directly using the graph edges, they encode graph adjacency information in the form of low-sensitivity cluster vectors, which are then perturbed using the Laplace mechanism to preserve edge-level privacy. Unlike our approach, however, neither of these methods provides node-level privacy guarantees.

Olatunji~\etal~\cite{olatunji2021releasing} propose the first node-level private GNN by adapting the framework of PATE~\cite{papernot2016semi}. In their approach, a student GNN model is trained on public graph data, with each node privately labeled using teacher GNN models that are trained exclusively for the corresponding query node. Nevertheless, their approach relies on public graph data and may not be applicable in all situations. Daigavane~\etal~\cite{daigavane2021node} extend the standard DP-SGD algorithm and privacy amplification by subsampling to bounded-degree graph data to achieve node-level DP, but their method fails to provide inference privacy. Finally, Sajadmanesh~\etal~\cite{sajadmanesh2023gap} propose \textsc{GAP}, a private GNN learning framework that provides both edge-level and node-level privacy guarantees using the aggregation perturbation approach. They decouple the aggregation steps from the neural network model to manage the privacy costs of their method. Although our method leverages the same aggregation perturbation technique, we take a different approach to limit the privacy costs using a progressive training scheme.

The main concept behind progressive learning is to train the model on simpler tasks first and then gradually move towards more challenging tasks. It was originally introduced to stabilize the training of deep learning models and has been widely adopted in various computer vision applications, such as facial attribute editing~\cite{wu2020cascade}, image super-resolution~\cite{wang2018fully}, image synthesis~\cite{karras2017progressive}, and representation learning~\cite{li2020progressive}. This technique has also been extended to federated learning, mainly to minimize the communication overhead between clients and the central server~\cite{pmlr-v119-belilovsky20a, he2021pipetransformer, wang2022progfed}. However, the potential benefit of progressive learning in DP applications has not been explored yet. In this paper, we are first to examine the advantages of progressive learning in the context of private GNNs.

\section{Background}\label{sec:background}

% This section begins by presenting key concepts and notations that will be used throughout the paper. Following this, we formally define the problem of training graph neural networks (GNNs) with differential privacy (DP).

\subsection{Differential Privacy}\label{sec:dp}

Differential privacy (DP) is a widely accepted framework for measuring the privacy guarantees of algorithms that operate on sensitive data. The main idea of DP is to ensure that the output of an algorithm is not significantly affected by the presence or absence of any particular individual's data in the input. This means that even if an attacker has access to all but one individual's data, they cannot determine whether that individual's data was used in the computation. The formal definition of DP is as follows~\cite{dwork2006calibrating}:
\begin{definition}\label{def:dp}
	Given $\epsilon > 0$ and $\delta\in[0,1]$, a randomized algorithm $\mathcal{A}$ satisfies $(\epsilon,\delta)$-differential privacy, if for all adjacent datasets $\mathcal{D}$ and $\mathcal{D}^\prime$ differing by at most one record and for all possible subsets of $\mathcal{A}$'s outputs $S\subseteq Range(\mathcal{A})$:
	\begin{equation*}
		\Pr[\mathcal{A}(\mathcal{D}) \in S] \le e^\epsilon\Pr[\mathcal{A}(\mathcal{D}^\prime) \in S] + \delta.
	\end{equation*}
\end{definition}

The value of epsilon, which is often called privacy budget or privacy cost parameter, determines the strength of the privacy guarantee provided by the algorithm, with smaller values of epsilon indicating stronger privacy protection but potentially lower utility of the algorithm. The parameter $\delta$ represents the maximum allowable failure probability, i.e., the probability that the algorithm may violate the privacy guarantee, and is usually set to a small value.

The guarantee of differential privacy depends on the notion of adjacency between datasets. In the case of tabular datasets, differential privacy defines adjacency between two datasets as being able to obtain one dataset from the other by removing (or replacing) a single record. However, for graph datasets, adjacency needs to be defined differently due to the presence of links between data records. 
To adapt the definition of DP for graphs, two different notions of adjacency are defined: edge-level and node-level adjacency. In the former, two graphs are adjacent if they differ only in the presence of a single edge, whereas in the latter, the two graphs differ by a single node with its features, labels, and all attached edges.
Accordingly, the definitions of edge-level and node-level DP are derived from these definitions~\cite{raskhodnikova2016differentially}. Specifically, an algorithm $\mathcal{A}$ provides edge-/node-level $(\epsilon, \delta)$-DP if for every two edge-/node-level adjacent graph datasets $\mathcal{G}$ and $\mathcal{G}^\prime$ and any set of outputs $S\subseteq Range(\mathcal{A})$, we have $\Pr[\mathcal{A}(\mathcal{G}) \in S] \le e^\epsilon\Pr[\mathcal{A}(\mathcal{G}^\prime) \in S] + \delta.$

The concepts of edge-level and node-level differential privacy can be intuitively understood as providing privacy protection at different levels of granularity in graph datasets. Edge-level differential privacy is focused on protecting the privacy of edges, which may represent connections between individuals. In contrast, node-level differential privacy aims to protect the privacy of nodes and their adjacent edges, thus safeguarding all information related to an individual, including their features, labels, and connections.

The difference in granularity between edge-level and node-level DP is crucial because the level of privacy protection needed may depend on the sensitivity of the information being disclosed. For example, protecting only the privacy of individual edges may be sufficient for some applications, while others may require more stringent privacy guarantees that protect the privacy of entire nodes. Our proposed method, which we discuss in detail in \autoref{sec:method}, is capable of providing both edge-level and node-level privacy guarantees.

\subsection{Graph Neural Networks}\label{sec:gnn}

The goal of GNNs is to learn a vector representation, also known as an embedding, for each node in a given graph. These embeddings are learned by taking into account the initial features of the nodes and the structure of the graph (i.e., its edges). The learned embeddings can be applied to various downstream machine learning tasks, such as node classification and link prediction. 

A common $K$-layer GNN is composed of $K$ layers of graph convolution that are applied sequentially. Specifically, layer $k$ takes as input the adjacency matrix $\mathbf{A}$ and the node embeddings produced by layer $k-1$, denoted by $\mathbf{X}^{(k-1)}$, and outputs a new embedding for each node by aggregating the embeddings of its adjacent neighbors, followed by a neural network transformation. In its simplest form, the formal update rule for layer $k$ can be written as follows:
\begin{equation}\label{eq:gnn-update-rule}
\mathbf{X}^{(k)} = \UPD\left(\AGG( \mathbf{A}, \mathbf{X}^{(k-1)} ); \mathbf{\Theta}^{(k)}\right),
\end{equation}
where $\AGG$ is a differentiable permutation-invariant \emph{neighborhood aggregation function}, such as mean, sum, or max pooling, and $\UPD$ denotes a learnable transformation, such as a multilayer perceptron (MLP), parameterized by $\mathbf{\Theta}^{(k)}$ that takes the aggregated embeddings as input and produces a new embedding for each node.
% , and can be adapted to incorporate edge features or learnable parameters. For instance, a simple sum aggregation function can be defined as:
% \begin{equation}
% \AGG\left( \mathbf{A}, \mathbf{X} \right) = \mathbf{A}^T\cdot\mathbf{X}.
% \end{equation}

% The input to the first layer is $\mathbf{X}^{(0)}=\mathbf{X}$, i.e., the initial node features. The output of the final layer $\mathbf{X}^{(K)}$ can then be used for downstream tasks, such as node classification or link prediction.

\subsection{Problem Definition}

Consistent with prior work~\cite{sajadmanesh2023gap,daigavane2021node}, we focus on the node classification task.
% which can be approached in two different problem settings: transductive and inductive. In the transductive setting, both training and testing are performed on the same graph, but with different sets of nodes used for each. Conversely, in the inductive setting, the training and testing are conducted on different graphs.
Let $\mathcal{G}=(\mathcal{V}, \mathcal{E})$ be a directed, unweighted graph with a set of nodes $\mathcal{V}=\{v_1,\dots,v_N\}$ and edges $\mathcal{E}$ represented by an adjacency matrix $\mathbf{A}\in\{0,1\}^{N\times N}$. 
The graph is associated with a set of node features and ground-truth labels. Node features are represented by a matrix $\mathbf{X}\in\mathbb{R}^{N\times d}$, where $\mathbf{X}_i$ denotes the $d$-dimensional feature vector of node $v_i$. Node labels are denoted by $\mathbf{Y}\in\{0,1\}^{N\times C}$, where $C$ is the number of classes, and $\mathbf{Y}_i$ is a one-hot vector indicating the label of node $v_i$. We assume that the node labels are known only for a subset of nodes $\mathcal{V}_L\subset\mathcal{V}$. This reflects the transductive (semi-supervised) learning setting, where the goal is to predict the labels of the remaining nodes in $\mathcal{V}\setminus\mathcal{V}_L$.

Consider a GNN-based node classification model $\mathcal{M}(\mathbf{A},\mathbf{X}; \mathbf{\Theta})$ with parameter set $\mathbf{\Theta}$ that takes the adjacency matrix $\mathbf{A}$ and the node features $\mathbf{X}$, and outputs the corresponding predicted node labels $\widehat{\mathbf{Y}}$:
\begin{equation}\label{eq:inference}
	\widehat{\mathbf{Y}} = \mathcal{M}(\mathbf{A},\mathbf{X}; \mathbf{\Theta}).
\end{equation}
We seek to minimize a standard classification loss function $\mathcal{L}$ with respect to the set of model parameters $\mathbf{\Theta}$ over the labeled nodes~$\mathcal{V}_L$:
\begin{align}\label{eq:train}
	\mathbf{\Theta}^\star   &= \arg\min_{\mathbf{\Theta}} \mathcal{L}(\widehat{\mathbf{Y}}, {\mathbf{Y}}) \nonumber \\
                          &= \arg\min_{\mathbf{\Theta}} \left( \sum_{v_i\in\mathcal{V}_L} \ell(\widehat{\mathbf{Y}}_i, \mathbf{Y}_i) \right),
\end{align}
where $\ell: \mathbb{R}^C \times \mathbb{R}^C \rightarrow \mathbb{R}$ is a loss function, such as cross-entropy, and $\mathbf{\Theta}^\star$ denotes the optimal set of parameters.

Our goal is to ensure the privacy of $\mathcal{G}$ at both the training (\autoref{eq:train}) and inference (\autoref{eq:inference}) phases of the model $\mathcal{M}$, using the differential privacy notions defined for graphs, i.e., edge-level and node-level DP. Note that preserving privacy during the inference stage is of utmost importance since the adjacency information of the graph is still used at inference time to generate the predicted labels, and thus sensitive information about the graph could potentially be leaked even with $\mathbf{\Theta}$ being differentially private~\cite{sajadmanesh2023gap}.

\section{Proposed Method}\label{sec:method}

\begin{figure}[t]
	\centering
	\includegraphics[width=\columnwidth]{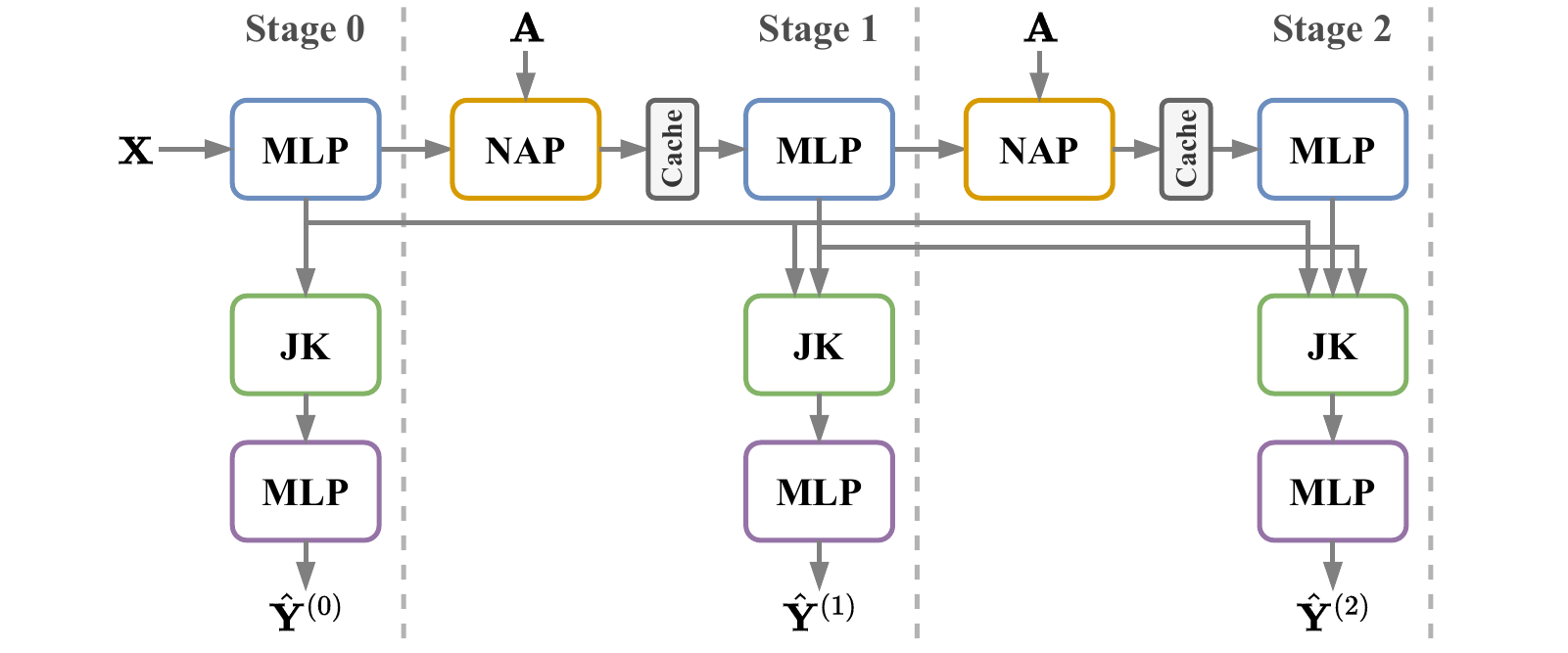}
	\caption{An example \name{} architecture with three stages (depth = 2). MLP and JK represent multi-layer perceptron and Jumping Knowledge~\cite{pmlr-v80-xu18c} modules, respectively. NAP denotes the normalize-aggregate-perturb module used to ensure the privacy of the adjacency matrix, with its output cached immediately after computation to save privacy budget. Training is done progressively, starting with the first stage and then expanding to the second and third stages, each using its own head MLP. The final prediction is obtained by the head MLP of the last stage.
	}
	\label{fig:architecture}
\end{figure}

In this section, we present our proposed \name{} method, which leverages the aggregation perturbation (AP) technique~\cite{sajadmanesh2023gap} to ensure differential privacy but introduces a novel progressive learning scheme to restrain the privacy costs of AP incurred during training. The overview of  \name{} architecture is illustrated in \autoref{fig:architecture}, and its forward propagation (inference) and training algorithms are presented in \autoref{alg:inference} and \autoref{alg:training}, respectively. In the following, we first describe our method in detail and then analyze its privacy guarantees. %The analysis of \name{}'s privacy guarantees are presented in \autoref{sec:privacy}.

\subsection{Model Architecture and Training}

\begin{algorithm}[t]
	\caption{\name{} Forward Propagation $\mathcal{M}_s\left( \mathbf{A}, \mathbf{X}; \sigma, \mathfrak{P}_s \right)$}\label{alg:inference}
	\small
	\SetKwInOut{Input}{Input}
	\SetKwInOut{Output}{Output}
	\ResetInOut{Output}
	\LinesNumbered
	\Input{Stage $s$, adjacency matrix $\mathbf{A}$; node features $\mathbf{X}$; noise standard deviation $\sigma$; model parameters $\mathfrak{P}_s = \bigcup_{k=0}^s\{\mathbf{\Theta}_{base}^{(k)}\}\cup\{\mathbf{\Theta}_{jump}^{(s)},\mathbf{\Theta}_{head}^{(s)}\}$
	}
	\Output{Predicated node labels $\widehat{\mathbf{Y}}^{(s)}$}
	\BlankLine
	\DontPrintSemicolon
	$\widetilde{\mathbf{X}}^{(0)} \gets \mathbf{X}$\;
	% if i greater than 0 then
	\For{$k\in\{0,\dots,s\}$} {
		\If{$k>0 \text{ and } \widetilde{\mathbf{X}}^{(k)} \text{ is not cached}$} {
			$\widetilde{\mathbf{X}}^{(k)} \gets \NAP(\mathbf{A}, {\mathbf{X}}^{(k-1)}; \sigma)$ \label{line:if}\;
			Cache $\widetilde{\mathbf{X}}^{(k)}$ \label{line:endif}\;
		}
		$\mathbf{X}^{(k)} \gets \MLP_{base}^{(k)}\left( \widetilde{\mathbf{X}}^{(k)}; \mathbf{\Theta}_{base}^{(k)} \right)$ \;
	}
	$\widehat{\mathbf{Y}}^{(s)} \gets \MLP_{head}^{(s)}\left( \text{JK}^{(s)}(\{ \mathbf{X}^{(0)},\dots,\mathbf{X}^{(s)} \}; \mathbf{\Theta}_{jump}^{(s)}); \mathbf{\Theta}_{head}^{(s)} \right)$\;
	\Return $\widehat{\mathbf{Y}}^{(s)}$\;
\end{algorithm}

\begin{algorithm}[t]
	\caption{\name{} Training}\label{alg:training}
	\small
	\SetKwInOut{Input}{Input}
	\SetKwInOut{Output}{Output}
	\ResetInOut{Output}
	\LinesNumbered
	\Input{Adjacency matrix $\mathbf{A}$; node features $\mathbf{X}$; node labels $\mathbf{Y}$; model depth $K$; noise standard deviation $\sigma$;}
	\Output{Trained model parameters $\mathfrak{P}^\star_{K}$}
	\BlankLine
	\DontPrintSemicolon
	% $\widetilde{\mathbf{X}}^{(0)} \gets \mathbf{X}$\;
	initialize $\mathbf{\Theta}_{base}^{(0)},\mathbf{\Theta}_{jump}^{(0)},\mathbf{\Theta}_{head}^{(0)}$ randomly\;
	$\mathfrak{P}_0 \gets \{\mathbf{\Theta}_{base}^{(0)},\mathbf{\Theta}_{jump}^{(0)},\mathbf{\Theta}_{head}^{(0)}\}$\;
	\For{$s\in\{0,\dots,K\}$} {
		$\mathfrak{P}^\star_s\gets\arg\min_{\mathfrak{P}}\mathcal{L}\Big(\mathcal{M}_{s}\left( \mathbf{A}, \mathbf{X}; \sigma, \mathfrak{P}_s \right), \mathbf{Y}\Big)$ \label{line:train}\;
		\If{$s<K$} {
			initialize $\mathbf{\Theta}_{base}^{(s+1)},\mathbf{\Theta}_{jump}^{(s+1)},\mathbf{\Theta}_{head}^{(s+1)}$ randomly\;
			$\mathfrak{P}_{s+1} \gets \mathfrak{P}^\star_{s} \cup \{\mathbf{\Theta}_{base}^{(s+1)},\mathbf{\Theta}_{jump}^{(s+1)},\mathbf{\Theta}_{head}^{(s+1)}\} \setminus \{\mathbf{\Theta^\star}_{jump}^{(s)},\mathbf{\Theta^\star}_{head}^{(s)}\}$ \;
		}
	}
	\Return $\mathfrak{P}^\star_{K}$\;
\end{algorithm}

% \subsection{The \name{} Model}
% Consider a non-private sequential graph neural network (GNN) model for node classification with $K$ aggregation steps (similar to \autoref{eq:gnn}) and an additional preprocessing layer as:
We start by considering a simple non-private sequential GNN model $\mathcal{M}$ with $K$ aggregation layers as the following:
\begin{align}
	\mathbf{X}^{(0)}     & = \MLP_{base}^{(0)}\left(\mathbf{X}; \mathbf{\Theta}_{base}^{(0)}\right),                                                                              \\
	\mathbf{X}^{(k)}     & = \MLP_{base}^{(k)}\left( \AGG( \mathbf{A}, \mathbf{X}^{(k-1)} ); \mathbf{\Theta}_{base}^{(k)} \right)\label{eq:gnn} \\
  & \quad \forall k\in\{1,\dots,K\}, \nonumber \\
	\widehat{\mathbf{Y}} & = \text{MLP}_{head}\left(\mathbf{X}^{(K)}; \mathbf{\Theta}_{head}\right),
\end{align}
where $\mathbf{X}^{(k)}$ is the node embeddings generated at layer $k$ by $\text{MLP}_{base}^{(k)}$ having parameters $\mathbf{\Theta}_{base}^{(k)}$, and $\text{MLP}_{head}$ is a multi-layer perceptron parameterized by $\mathbf{\Theta}_{head}$ with the softmax activation function that maps the final embeddings $\mathbf{X}^{(K)}$ to the predicted class probabilities $\widehat{\mathbf{Y}}$.

To make this model differentially private, we follow the aggregation perturbation technique proposed by Sajadmanesh\etal~\cite{sajadmanesh2023gap} and add noise to the output of the aggregation function. Specifically, we replace the original aggregation function $\AGG$ in \autoref{eq:gnn} with a \emph{Normalize-Aggregate-Perturb} mechanism defined as:
\begin{gather}
  \NAP\left(\mathbf{A}, \mathbf{X}; \sigma\right) = \nonumber \\
  \left[ \sum_{j=1}^N \frac{\mathbf{X}_j}{\norm[2]{\mathbf{X}_j}}\mathbf{A}_{j,i} + \mathcal{N}(\mathbf{0}, \sigma^2\mathbf{I}_d) \mid \forall i\in\{1,\dots,N\} \right],
\end{gather}
% where
% \begin{equation}
% 	\widetilde{\mathbf{X}}_i = \sum_{j=1}^N \mathbf{A}_{j,i}\frac{\mathbf{X}_i}{\norm[2]{\mathbf{X}_i}} + \mathcal{N}(\mathbf{0}, \sigma^2\mathbf{I}_d) \mid \forall i\in\{1,\dots,N\},
% \end{equation}
where $N$ is the number of nodes, $d$ is the dimension of the input node embeddings, and $\sigma$ is the standard deviation of the Gaussian noise. Concretely, the NAP mechanism row-normalizes the input embeddings to limit the contribution of each node to the aggregated output, then applies the sum aggregation function followed by adding Gaussian noise to the results.

It can be easily shown that the resulting model provides edge-level DP as every query to the adjacency matrix $\mathbf{A}$ is immediately perturbed with noise.
%For the node-level DP, however, the additional requirement is the use of DP-SGD to further protect node features and labels.
However, training such a model comes at the cost of a significant increase in the privacy budget, which is proportional to the number of queries to the adjacency matrix.
Concretely, with $T$ training iterations, the \NAP{} mechanism is queried $KT$ times (at each forward pass and each layer), leading to an excessive accumulated privacy cost of $O(\sqrt{KT})$.

To reduce this cost, we propose a progressive training approach as the following: We first split the model $\mathcal{M}$ into $K+1$ overlapping submodels, where submodel $\mathcal{M}_s$, $s\in\{0,1,\dots, K\}$, is defined as:
% \todo{Can we merge (8) and (9) into one equation?}
\begin{align}
	\widetilde{\mathbf{X}}^{(s)} & = \NAP\left( \mathbf{A}, {\mathbf{X}}^{(s-1)}; \sigma \right),\label{eq:nap}                                                       \\
	\mathbf{X}^{(s)}             & = \MLP_{base}^{(s)}\left( \widetilde{\mathbf{X}}^{(s)}; \mathbf{\Theta}_{base}^{(s)} \right),                                   \\
	\widehat{\mathbf{Y}}^{(s)} &= \MLP_{head}^{(s)}\left( \text{JK}^{(s)}(\bigcup_{k=0}^{s} \{\mathbf{X}^{(k)} \}; \mathbf{\Theta}_{jump}^{(s)}); \mathbf{\Theta}_{head}^{(s)} \right),
\end{align}
where $\widetilde{\mathbf{X}}^{(s)}$ is the noisy aggregate embeddings of $\mathcal{M}_s$, with $\widetilde{\mathbf{X}}^{(0)} = \mathbf{X}$. $\text{JK}^{(s)}$ is a Jumping Knowledge module~\cite{pmlr-v80-xu18c} with parameters $\mathbf{\Theta}_{jump}^{(s)}$ that combines the embeddings generated by submodels $\mathcal{M}_0$ to $\mathcal{M}_s$, and $\MLP^{(s)}_{head}$ is a lightweight, 1-layer head MLP with parameters $\mathbf{\Theta}_{head}^{(s)}$ used to train $\mathcal{M}_s$. Finally, $\widehat{\mathbf{Y}}^{(s)}$ is the output predictions of $\mathcal{M}_s$.
%With each submodel equipped with its own head MLP, 
Then, we progressively train the model in $K+1$ stages, starting from the shallowest submodel $\mathcal{M}_0$ and gradually expanding to the deepest submodel $\mathcal{M}_K$ (which is equivalent to the full model $\mathcal{M}$) as explained by \autoref{alg:training}. For the final inference after training, we simply use the labels predicted by the last submodel $\mathcal{M}_K$, i.e., $\widehat{\mathbf{Y}}=\widehat{\mathbf{Y}}^{(K)}$.
%More formally, at each stage $s\in\{0,1,\dots,K\}$, we train $\mathcal{M}_s$ using the loss function $\mathcal{L}^{(k)}=\mathcal{L}(\widehat{\mathbf{Y}}^{(k)}, \mathbf{Y}; \mathbf{\Theta}^{(k)})$ that is computed using the predictions $\widehat{\mathbf{Y}}^{(k)}$ and optimized over the parameters $\mathbf{\Theta}^{(k)}$ of submodel $s$.

\emph{The key point in this training strategy is that we immediately save the outputs of NAP modules on their first query and reuse them throughout the training.} More specifically, at each stage $s$, the perturbed aggregate embedding matrix $\widetilde{\mathbf{X}}^{(s)}$ computed in the first forward pass of $\mathcal{M}_s$ (via \autoref{eq:nap}) is stored in the cache and reused in all further queries. This caching mechanism allows us to reduce the privacy costs of the model by a factor of $T$, as the NAP module in this case is only queried $K$ times (once per stage) instead of $KT$ times. At the same time, the aggregations $\widetilde{\mathbf{X}}^{(s)}$ are computed over the embeddings $\mathbf{X}^{(s-1)}$ that are already learned in the preceding stage $s-1$, which provide more expressive power than the raw node features as they also encode information from the adjacency matrix and node labels, and thus lead to better performance.

% \begin{remark}
%   The proposed \name{} model can also be trained in a layerwise fashion, i.e., by training each layer $\MLP_{base}^{(k)}$ individually, while keeping the parameters of the preceding layers frozen and using the same caching mechanism. Note that this is slightly different from the progressive approach, in which all the parameters from layer 0 to layer $s$, i.e., $\mathbf{\Theta}_{base}^{(0)},\dots,\mathbf{\Theta}_{base}^{(s)}$ are trained together in each stage $s$. In \autoref{sec:results}, we show that such a progressive training strategy leads to better performance than layerwise training.
% \end{remark}

% \todo[inline]{add remark: progressive -> transductive, layerwise -> inductive}

\subsection{Privacy Analysis}

With the following theorem, we show that the proposed training strategy provides edge-level DP. The proof is provided in 
the supplementary material.
% Appendix~\ref{sec:proof:privacy}.

\begin{theorem}\label{thm:privacy}
	Given the model depth $K\ge0$ and noise variance $\sigma^2$, for any $\delta\in(0,1)$ \autoref{alg:training} satisfies edge-level $(\epsilon,\delta)$-DP with $\epsilon=\frac{K}{2\sigma^2} + \nicefrac{\sqrt{2K\log{(1/\delta)}}}{\sigma}$.
\end{theorem}

To ensure node-level DP, however, we must train every submodel using DP-SGD or its variants, as in this case node features and labels are also private and can be leaked with non-private training. \autoref{thm:nodeprivacy} establishes the node-level DP guarantee of \name{}'s training algorithm when combined with DP-SGD:
\begin{theorem}\label{thm:nodeprivacy}
	Given the number of nodes $N$, batch-size $B<N$, number of per-stage training iterations $T$, gradient clipping threshold $C>0$, model depth $K\ge0$, maximum cut-off degree $D\ge1$, noise variance for aggregation perturbation $\sigma^2_{AP}>0$, and noise variance for gradient perturbation $\sigma^2_{GP}>0$, \autoref{alg:training} satisfies node-level $(\epsilon, \delta)$-DP for any $\delta\in(0,1)$ with:
      \begin{multline}
        \epsilon \le \min_{\alpha>1} \frac{(K+1)T}{\alpha-1}\log\left\lbrace \left(1-\frac{B}{N}\right)^{\alpha-1}\left( \alpha\frac{B}{N} - \frac{B}{N} + 1 \right) \right. \\
         \left. + \binom{\alpha}{2}\left(\frac{B}{N}\right)^2\left(1-\frac{B}{N}\right)^{\alpha-2}e^{\frac{C^2}{\sigma_{GP}^2}} \right. \\
         \left. +\sum_{l=3}^\alpha\binom{\alpha}{l}\left(1-\frac{B}{N}\right)^{\alpha-l}\left(\frac{B}{N}\right)^l e^{(l-1)(\frac{C^2 l}{2\sigma_{GP}^2})} \right\rbrace \\
         + \frac{DK\alpha}{2\sigma_{AP}^2} + \frac{\log(1/\delta)}{\alpha-1},
      \end{multline}
providing that the optimization in \autoref{line:train} of \autoref{alg:training} is done using DP-SGD.
\end{theorem}

The proof is available in 
the supplementary material.
% Appendix~\ref{sec:proof:nodeprivacy}. 
Note that to decrease the node-level sensitivity of the NAP mechanism (i.e., the impact of adding/removing a node on the output of the NAP mechanism), we assume an upper bound $D$ on node degrees, and randomly sample edges from the graph to ensure that each node has no more than $D$ outgoing edges. This is a standard technique to ensure bounded-degree graphs~\cite{daigavane2021node,sajadmanesh2023gap}.

In addition to training privacy, \name{} also guarantees privacy during inference at both edge and node levels without any further privacy costs. This is because the entire noisy aggregate matrices $\widetilde{\mathbf{X}}^{(k)}$ corresponding to all the nodes --both training and test ones-- are already computed and cached during training and reused for inference (i.e., lines \ref{line:if} and \ref{line:endif} of \autoref{alg:inference} is not executed at inference time). 
Therefore, the inference for a node $i$ no longer depends on the adjacency matrix and neighboring node features, but only on its own aggregated features $\widetilde{\mathbf{X}}^{(k)}_i$ for all $k\in\{0,\dots,K\}$, which are already computed with DP during training. As a result, the inference only post-processes differentially private outputs, which does not incur any additional privacy costs.

% \todo[inline]{refer to the new paper for the definition of inference privacy}
\section{Experimental Setup}\label{sec:setup}

We test our proposed method on node-wise classification tasks and evaluate its effectiveness in terms of classification accuracy and privacy guarantees.

\subsection{Datasets}
We conduct experiments on three real-world datasets that have been used in previous work~\cite{sajadmanesh2023gap,daigavane2021node,olatunji2021releasing}, namely Facebook~\cite{traud2012social}, Reddit~\cite{hamilton2017inductive}, and Amazon~\cite{chiang2019cluster}, and also two new datasets: FB-100~\cite{traud2012social} and WeNet~\cite{giunchiglia2021worldwide, meegahapola2023generalization}. The Facebook dataset is a collection of anonymized social network data from UIUC students, where nodes represent users, edges indicate friendships, and the task is to predict students' class year. 
The Reddit dataset comprises a set of Reddit posts as nodes, where edges represent if the same user commented on both posts, and the goal is to predict the posts' subreddit. 
The Amazon dataset is a product co-purchasing network, with nodes representing products and edges indicating if two products are purchased together, and the objective is to predict product category.
% In addition to the above datasets, we use two new datasets: FB-100~\cite{traud2012social} and WeNet~\cite{giunchiglia2021worldwide, meegahapola2023generalization}. 
FB-100 is an extended version of the Facebook dataset combining the social network of 100 different American universities. WeNet is a mobile sensing dataset collected from university students in four different countries. Nodes represent eating events, which are linked based on the similarity of location and Wi-Fi sensor readings. Node features are extracted based on cellular and application sensors, and the goal is to predict the country of the events.%\footnote{Unlike the other four datasets, WeNet is not publicly available.} 
A summary of the datasets is provided in \autoref{tab:datasets}.

\begin{table}[t]
  \centering
  \caption{Dataset Statistics}
  \label{tab:datasets}
  \sc
  \small
  \SetTblrInner{colsep=2.5pt}
  \begin{tblr}{Q[l,m]X[c,m]X[c,m]Q[c,m]Q[c,m]X[c,m]}
    \toprule
        Dataset &     \# Nodes &     \# Edges & \# Features & \# Classes & {Degree} \\
    \midrule
    Facebook &    26,406 &  2,117,924 &      501 &       6 &            62 \\
    Reddit &   116,713 & 46,233,380 &      602 &       8 &           209 \\
    Amazon & 1,790,731 & 80,966,832 &      100 &      10 &            22 \\
FB-100 & 1,120,280 & 86,304,478 &      537 &       6 &            57 \\
     WeNet &    37,576 & 22,684,206 &       44 &       4 &           286 \\
    \bottomrule
  \end{tblr}
\end{table}

\begin{table*}[t]
	\centering
	\caption{Comparison of Experimental Results (Mean Accuracy $\pm$ 95\% CI)
  }
	\label{tab:results}
	\sc
	\small
	% \SetTblrInner{colsep=3pt}
	\begin{tblr}{Q[l,m]Q[l,m]Q[c,m]Q[c,m]Q[c,m]Q[c,m]Q[c,m]Q[c,m]}
		\toprule
		Privacy Level & Method & $\epsilon$ &  Facebook & Reddit & Amazon & FB-100 & WeNet \\
    \midrule
		\SetCell[r=3]{m} {Non-Private}
    & GraphSAGE~\cite{hamilton2017inductive} & $\infty$ & \textbf{84.7 $\pm$ 0.09} & 99.4 $\pm$ 0.01 & 93.2 $\pm$ 0.07 & 74.0 $\pm$ 0.80 & 71.6 $\pm$ 0.54 \\
		& GAP~\cite{sajadmanesh2023gap}    & $\infty$ & 80.5 $\pm$ 0.42 & \textbf{99.5 $\pm$ 0.01} & 92.0 $\pm$ 0.10 & 66.4 $\pm$ 0.35 & 69.7 $\pm$ 0.14 \\
    & \name{} (\textbf{Ours}) & $\infty$ & {84.5 $\pm$ 0.24} & 99.3 $\pm$ 0.03 & \textbf{93.3 $\pm$ 0.04} & \textbf{74.4 $\pm$ 0.14} & \textbf{73.9 $\pm$ 0.25} \\
    \midrule
		\SetCell[r=4]{m} {Edge-Level\\Private} 
    & MLP & 0.0      & 50.8 $\pm$ 0.20 & 82.5 $\pm$ 0.08 & 71.1 $\pm$ 0.18 & 34.9 $\pm$ 0.02 & 51.5 $\pm$ 0.22 \\
    & EdgeRand~\cite{wu2021linkteller} & 1.0      & 50.2 $\pm$ 0.50 & 82.8 $\pm$ 0.05 & 72.7 $\pm$ 0.1 & 34.9 $\pm$ 0.05 & 52.1 $\pm$ 0.48 \\
    % & LPGNet~\cite{kolluri2022lpgnet} & 1.0      & 72.9 $\pm$ 0.20 & OOM & OOM & OOM & 28.6 $\pm$ 0.16 \\
    & GAP~\cite{sajadmanesh2023gap}    & 1.0      & 69.4 $\pm$ 0.39 & 97.5 $\pm$ 0.06 & 78.8 $\pm$ 0.26 & 46.5 $\pm$ 0.58 & 62.4 $\pm$ 0.28 \\
    & \name{} (\textbf{Ours}) & 1.0      & \textbf{77.2 $\pm$ 0.33} & \textbf{97.8 $\pm$ 0.05} & \textbf{84.2 $\pm$ 0.07} & \textbf{56.9 $\pm$ 0.30} & \textbf{68.8 $\pm$ 0.23} \\
    \midrule
		\SetCell[r=4]{m} {Node-Level\\Private} 
    & DP-MLP~\cite{abadi2016deep} & 8.0      & 50.2 $\pm$ 0.25 & 81.5 $\pm$ 0.12 & 73.6 $\pm$ 0.05 & 34.5 $\pm$ 0.13 & 50.8 $\pm$ 0.37  \\
    & DP-GNN~\cite{daigavane2021node} & 8.0      & 62.6 $\pm$ 0.86 & \textbf{95.6 $\pm$ 0.31} & 78.5 $\pm$ 0.86 & 46.5 $\pm$ 0.57 & 54.2 $\pm$ 0.73 \\
    & GAP~\cite{sajadmanesh2023gap}    & 8.0      & 63.9 $\pm$ 0.49 & 93.9 $\pm$ 0.09 & 77.6 $\pm$ 0.07 & 43.0 $\pm$ 0.20 & 58.2 $\pm$ 0.39 \\
    & \name{} (\textbf{Ours}) & 8.0      & \textbf{69.3 $\pm$ 0.33} & 94.0 $\pm$ 0.04 & \textbf{79.1 $\pm$ 0.10} & \textbf{48.5 $\pm$ 0.36} & \textbf{61.0 $\pm$ 0.34} \\
		\bottomrule
	\end{tblr}
\end{table*}

\subsection{Baselines}
We compare \name{} against the following baselines: 

\begin{itemize}[leftmargin=0pt, label={}]
  \item \textbf{\textsc{GraphSAGE}~\cite{hamilton2017inductive}.} This is one of the most popular GNN models, which we use for non-private performance comparison with our method. Moreover, it serves as the backbone model for the following \textsc{EdgeRand} and \textsc{DP-GNN} baselines. The number of message-passing layers is tuned in the range of 1 to 5 for each dataset. We also use one preprocessing and one postprocessing layer, and equip the model with jumping knowledge modules~\cite{pmlr-v80-xu18c} to get better performance.

  \item \textbf{MLP and DP-MLP.} We use a simple 3-layer MLP model which does not use any graph structural information and therefore is perfectly edge-level private. DP-MLP is the node-level private variant of MLP, which is trained using the DP version of the Adam optimizer (DP-Adam).
  
  \item \textbf{\textsc{EdgeRand}~\cite{wu2021linkteller}.} This edge-level private method directly adds noise to the adjacency matrix. We use the enhanced version of \textsc{EdgeRand} from~\cite{sajadmanesh2023gap} that uses the Asymmetric Randomized Response~\cite{imola2021comm} with the GraphSAGE model as the backbone GNN.
  
  % \item \textbf{\textsc{LPGNet}~\cite{kolluri2022lpgnet}.} We utilize LPGNet as another edge-level DP baseline. We employ its official implementation\footnote{\url{https://github.com/aashishkolluri/lpgnet-prototype}} with the default set of hyperparameters.
  
  \item \textbf{DP-GNN~\cite{daigavane2021node}}. This node-level private approach extends the DP-SGD algorithm to GNNs. Note, however, that this method does not support inference privacy, but we nevertheless include it in our comparison.
  Similar to \EdgeRand, we use the \GraphSAGE model as the backbone GNN for this method as well. %While the method supports for multiple message-passing layers, we did not observe any improvement in performance beyond one layer, and thus we only use one message-passing layer for all the datasets.
  
  \item \textbf{GAP~\cite{sajadmanesh2023gap}.} GAP is the state-of-the-art approach that supports both edge-level and node-level DP. We use GAP's official implementation on GitHub\footnote{\url{https://github.com/sisaman/GAP}} and follow the same experimental setup as reported in the original paper.
\end{itemize}

We do not include other available differentially private GNN baselines (e.g., \cite{sajadmanesh2021locally, olatunji2021membership, mueller2022differentially,ayle2023training}) as they have different problem settings that make them not directly comparable to our method.

\subsection{Implementation Details} 
We follow the same experimental setup as GAP~\cite{sajadmanesh2023gap}, and randomly split the nodes in all the datasets into training, validation, and test sets with 75/10/15\% ratio, respectively. We vary $\epsilon$ within $\{0.25,0.5,1,2,4,\infty\}$ for the edge-level privacy ($\epsilon=\infty$ corresponds to the non-private setting) and within $\{2,4,8,16,32\}$ for the node-level privacy setting. For each $\epsilon$ value, we tune the following hyperparameters based on the mean validation set accuracy computed over 10 runs: $\MLP_{base}$ layers in $\{1,2\}$, model depth $K$ in $\{1,2,3,4,5\}$, and learning rate in $\{0.01,0.05\}$. The value of $\delta$ is fixed per each dataset to be smaller than the inverse number of private units (i.e., edges for edge-level privacy, nodes for node-level privacy). For all cases, we set the number of $\MLP_{head}$ layers to 1 and use concatenation for the JK modules. Additionally, we set the number of hidden units to 16 and use the SeLU activation function~\cite{klambauer2017self}. We use batch normalization except for the node-level setting, for which we use group normalization with one group. Under the edge-level setting, we train the models with full-sized batches for 100 epochs using the Adam optimizer and perform early stopping based on the validation set accuracy. For the node-level setting, we use randomized neighbor sampling to bound the maximum degree $D$ to 50 for Amazon, 100 for Facebook and FB-100, and 400 for Reddit and WeNet. We use DP-Adam~\cite{gylberth2017differentially} with a clipping threshold of 1.0. We tune the number of per-stage epochs in $\{5,10\}$ and set the batch size to 256, 1024, 2048, 4096, and 4096 for Facebook, WeNet, Reddit, Amazon, and FB-100, respectively. Finally, we report the average test accuracy over 10 runs with 95\% confidence intervals calculated by bootstrapping with 1000 samples. 
We open-source our implementation on GitHub.\footnote{
\url{https://github.com/sisaman/ProGAP}
% It will be made public upon acceptance.
}

% \todo[inline]{describe hardware}
\subsection{Software and Hardware}
We use PyTorch Geometric~\cite{Fey/Lenssen/2019} for implementing the models, \texttt{autodp}\footnote{\url{https://github.com/yuxiangw/autodp}} for privacy accounting, and Opacus~\cite{opacus} for DP training. We run all the experiments on an HPC cluster with NVIDIA Tesla V100 and GeForce RTX 3090 GPUs with maximum 32GB memory.

\section{Results and Discussion}\label{sec:results}

\begin{figure*}[t]
	\centering
	\includegraphics[width=.8\textwidth]{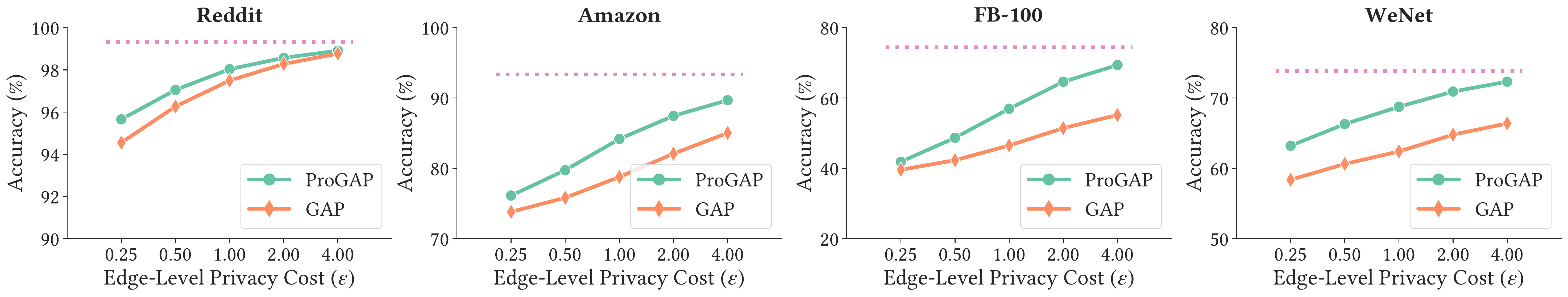}\\
  \includegraphics[width=.8\textwidth]{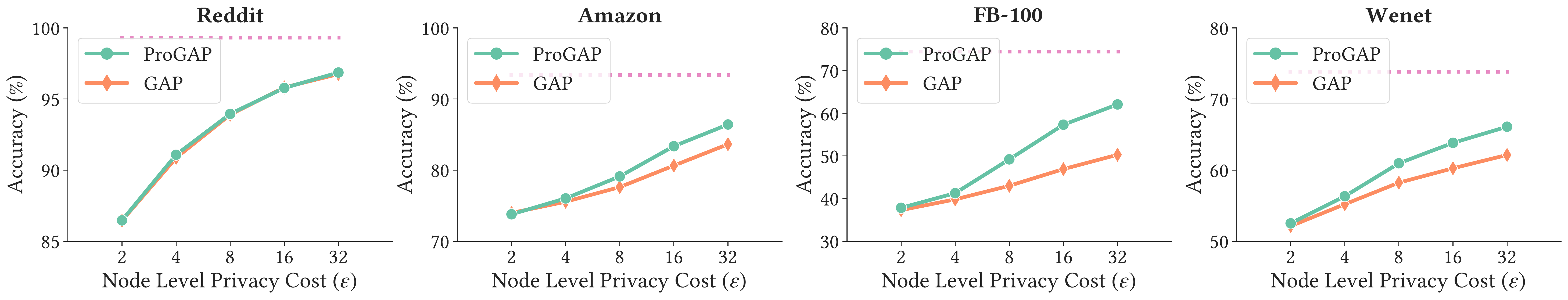}
	\caption{Accuracy-privacy trade-off of edge-level (top) and node-level (bottom) private methods. The dotted line represents the accuracy of the non-private \name{}.}
	\label{fig:acc-priv}
\end{figure*}

\subsection{Accuracy-Privacy Trade-off}\label{sec:acc-priv}

\autoref{tab:results} presents the test accuracy of \name{} against other baselines at three different privacy levels: non-private with $\epsilon=\infty$, edge-level privacy with $\epsilon=1$, and node-level privacy with $\epsilon=8$. The results are reported as mean accuracy $\pm$ 95\% confidence interval. We observe that \name{} outperforms other approaches most of the time, and often by a substantial margin. In the non-private setting, \name{} performs comparably with \GraphSAGE{}, but achieves higher test accuracies than GAP on all datasets except Reddit, where GAP performs only slightly better. This shows that our \name{} method is also a strong predictor in the non-private learning setting. 

Moving to edge-level privacy, \name{} consistently outperforms other approaches across all datasets, with the largest performance gap of 10.4\% accuracy points compared to GAP observed on FB-100. 
% Notably, \LPGNet{} suffers from out-of-memory (OOM) issues on the larger datasets, namely Reddit, Amazon, and FB-100, while our \name{} method scales well to large graphs.
Under node-level DP, \name{} is still superior to the other baselines across all datasets except Reddit, where DP-GNN achieves a slightly higher accuracy. Note, however, that DP-GNN only guarantees node-level privacy for the model parameters and fails to provide inference privacy, while our \name{} method provides both training and inference privacy guarantees. Compared to GAP, the largest margin in this category is also observed on FB-100, where \name{} achieves 5.5\% more accuracy points.

To examine the performance of \name{} at different privacy budgets, we varied $\epsilon$ between 0.25 to 4 for edge-level privacy and 2 to 32 for node-level private algorithms. We then recorded the accuracy of \name{} for each privacy budget and compare it with GAP as the most similar baseline. The outcome for both edge-level and node-level privacy settings is depicted in \autoref{fig:acc-priv}. %\footnote{The results on the Facebook dataset are omitted in the interest of the legibility of the plots.} 
Notably, we observe that \name{} achieves higher accuracies than GAP across all $\epsilon$ values tested and approaches the non-private accuracy more quickly under both privacy settings. This is because in \name{}, each aggregation step is computed on the node embeddings learned in the previous stage, providing greater representational power than GAP, which just recursively computes the aggregations on the initial node representations. 

It is worth noting that the performance discrepancy between ProGAP and GAP is not consistent across all datasets. 
For instance, this gap in accuracy is less pronounced with the Reddit dataset compared to FB-100. This is due to the specific characteristics and the learning task of each dataset, which require different levels of graph representational power. In Reddit, where the goal is to predict the community of nodes (representing Reddit posts), most of the pertinent information needed is already present in the node features, making the relationships between the posts less crucial for this prediction task. In contrast, the learning task of the FB-100 dataset (predicting students' class year) relies more heavily on the graph structure, necessitating more powerful graph representations. Therefore, the performance difference between \name{} and GAP is more noticeable in this dataset. This connection between the learning task and the graph structure will be revisited in \autoref{sec:ablation}.

\begin{figure*}[t]
	\centering
	\includegraphics[width=.8\textwidth]{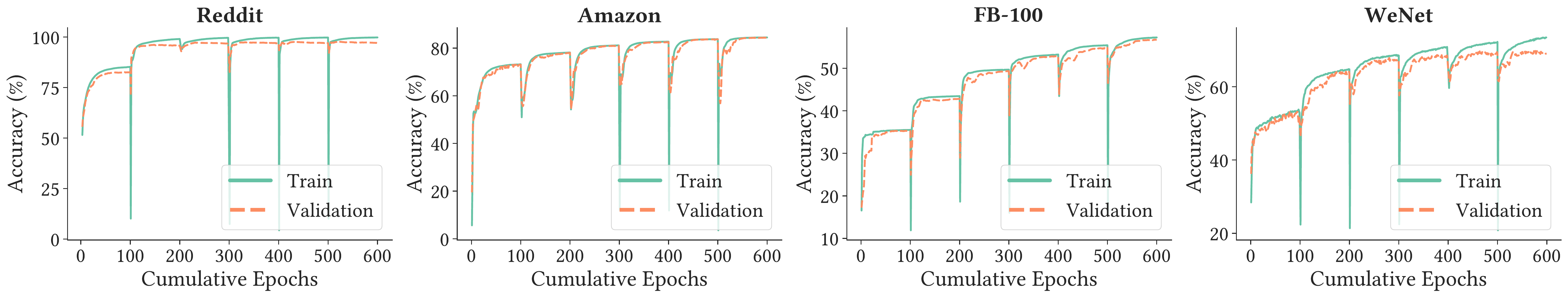}\\
  \includegraphics[width=.8\textwidth]{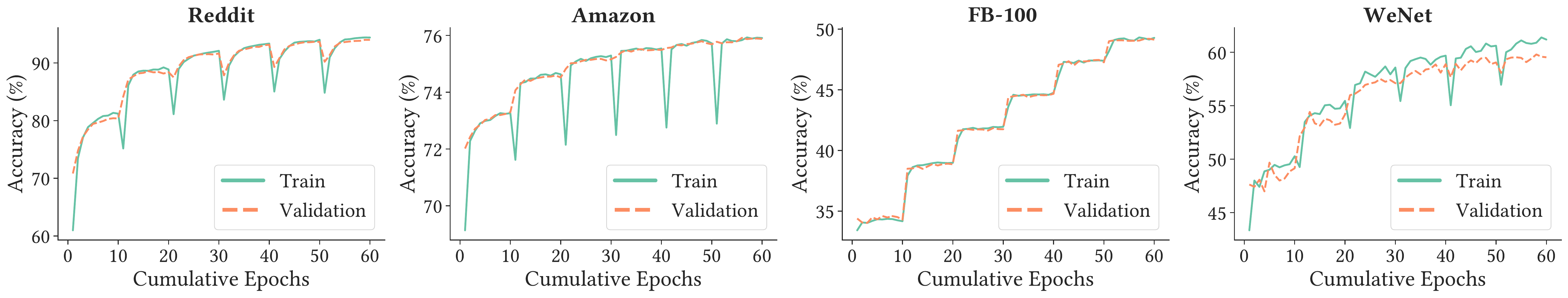}
	\caption{Convergence of \name{} with $K=5$ under edge-level (top) and node-level (bottom) privacy, with $\epsilon=1$ and $\epsilon=8$, respectively.}
	\label{fig:convergence}
\end{figure*}

\subsection{Convergence Analysis}
We examine the convergence of \name{} to further understand its behavior under the two privacy settings. We report the training and validation accuracy of \name{} per training step under edge-level privacy with $\epsilon=1$ and node-level privacy with $\epsilon=8$. For all datasets, \name{} is trained for 100 and 10 epochs per stage under edge and node-level privacy, respectively. We fix $K=5$ in all settings. The results are shown in \autoref{fig:convergence}. We observe that both training and validation accuracies increase as \name{} moves from stage 0 to 5, with diminishing returns for more stages, which indicates the higher importance of the nearby neighbors to each node, since the receptive field of nodes grows with the number of stages. 
% The convergence is faster under edge-level privacy, which is expected since the amount of noise injected during training is smaller than in node-level privacy. 
Moreover, we observe negligible discrepancies between training and validation accuracy when the model converges, which suggests higher resilience to privacy attacks, such as membership inference, which typically rely on large generalization gaps. This result is in line with previous work showing the effectiveness of DP against privacy attacks~\cite{jagielski2020auditing,jayaraman2019evaluating,nasr2021adversary,sajadmanesh2023gap}.

\begin{figure*}[t]
	\centering
	\includegraphics[width=.8\textwidth]{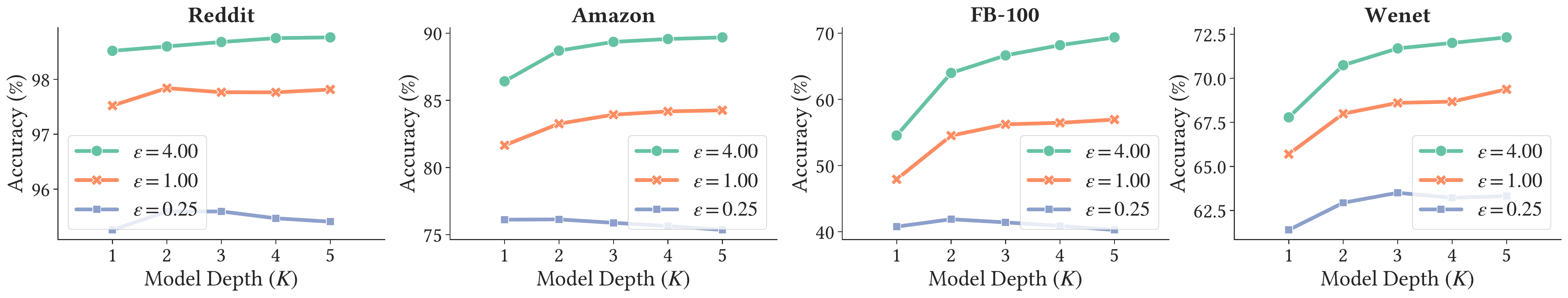}\\
  \includegraphics[width=.8\textwidth]{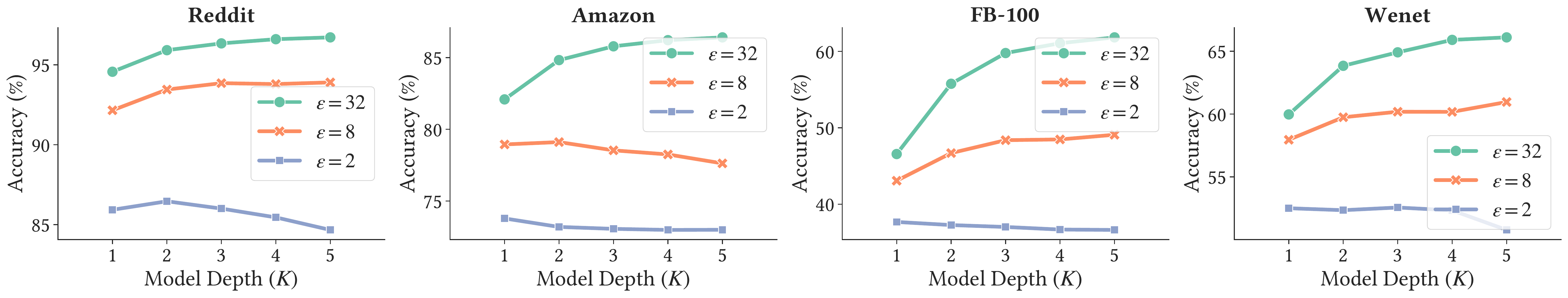}
	\caption{Effect of the model depth on \name{}'s accuracy under edge-level (top) and node-level (bottom) privacy.}
	\label{fig:depth}
\end{figure*}

\subsection{Effect of the Model Depth}\label{sec:ablation}
We explore how the performance of \name{} is influenced by modifying the model depth $K$, or equivalently, the number of stages $K+1$. We experiment with different values of $K$ ranging from 1 to 5 and evaluate \name{}'s accuracy under varying privacy budgets of $\epsilon\in\{0.25,1,4\}$ for edge-level DP and $\epsilon\in\{2, 8, 32\}$ for node-level privacy. The results are demonstrated in \autoref{fig:depth}. We observe that \name{} can generally gain advantages from increasing the depth, but there is a compromise depending on the privacy budget: 
deeper models lead to better accuracy under higher privacy budgets, while lower privacy budgets require shallower models to achieve optimal performance.
This is because \name{} can leverage data from more remote nodes with a higher value of $K$, which can boost the final accuracy, but it also increases the amount of noise in the aggregations, which has a detrimental effect on the model's accuracy. When the privacy budget is lower and the amount of noise is greater, \name{} has the best performance at smaller values of $K$. But as the privacy budget grows, the magnitude of the noise is lowered, enabling the model to take advantage of greater $K$ values.

An intriguing aspect to note is the potential improvement in \name{}'s accuracy across various datasets as its depth increases. Observing the Reddit and FB-100 datasets as an example, it is clear that the boost in performance from increasing the parameter $K$ is considerable for FB-100 under moderate privacy budgets, while for Reddit, the gain in accuracy is minimal.
This discrepancy can again be explained by the nature and the specific learning tasks of the datasets. In Reddit, where the task is less reliant on the graph structure and the node features hold more predictive value, the depth of \name{} doesn't significantly influence its performance. On the contrary, in FB-100, where the learning task heavily depends on the graph structure, \name{}'s performance is more sensitive to the model's depth. These observations align with the previous discussion about the wider performance gap between \name{} and GAP in FB-100 compared to Reddit.

\section{Conclusion}\label{sec:conclusion}

In this paper, we introduced \name{}, a novel differentially private GNN that improves the challenging accuracy-privacy trade-off in learning from graph data. Our approach uses a progressive training scheme that splits the GNN into a sequence of overlapping submodels, each of which is trained over privately aggregated node embeddings learned and cached by the previous submodels. By combining this technique with the aggregation perturbation method, we formally proved that \name{} can ensure edge-level and node-level privacy guarantees for both training and inference stages. Empirical evaluations on benchmark graph datasets demonstrated that \name{} can achieve state-of-the-art accuracy by outperforming existing methods. Future work could include exploring new architectures or training strategies to further improve the accuracy-privacy trade-off of differentially private GNNs, especially in the more challenging node-level privacy setting.

\begin{acks}
This work was supported by the European Commission's H2020
Program ICT-48-2020, AI4Media Project, under grant number 951911.
It was also supported by the European Commission's H2020 WeNet Project,
under grant number 823783.
\end{acks}

\bibliographystyle{ACM-Reference-Format}
\bibliography{ref}

\clearpage
\appendix
\subtitle{Supplementary Material}
\maketitlesup
% \hypersetup{hidelinks}
% \color{white}

\section{Rényi Differential Privacy}\label{sec:rdp}

The proofs presented in the following section are based on \emph{Rényi Differential Privacy} (RDP)~\cite{mironov2017renyi}, which is an alternative definition of DP that gives tighter sequential composition results. The formal definition of RDP is as follows:
\begin{definition}[Rényi Differential Privacy~\cite{mironov2017renyi}]
	Given $\alpha > 1$ and $\epsilon > 0$, a randomized algorithm $\mathcal{A}$ satisfies $(\alpha,\epsilon)$-RDP if for every adjacent datasets $X$ and $X^\prime$, we have:
  \begin{equation}
    D_\alpha\left( \mathcal{A}(\mathcal{D}) \Vert \mathcal{A}(\mathcal{D}^\prime) \right) \le \epsilon,
  \end{equation}
	where $D_\alpha(P\Vert Q)$ is the {Rényi divergence} of order $\alpha$ between probability distributions $P$ and $Q$ defined as:
	\begin{equation*}
		D_\alpha(P\Vert Q) = \frac{1}{\alpha-1}\log\mathbb{E}_{x\sim Q}\left[\frac{P(x)}{Q(x)}\right]^\alpha.
	\end{equation*}
\end{definition}

Proposition~3 of~\cite{mironov2017renyi} measures the privacy guarantee of the composition of RDP algorithms as the following:
\begin{proposition}
  [Composition of RDP mechanisms~\cite{mironov2017renyi}]
  \label{prop:rdpcompose}
    Let $f$ be $(\alpha,\epsilon_1)$-RDP and $g$ be $(\alpha,\epsilon_2)$-RDP, then the mechanism defined as $(X,Y)$, where $X\sim f(\mathcal{D})$ and $Y\sim g(X,\mathcal{D})$, satisfies $(\alpha,\epsilon_1+\epsilon_2)$-RDP.
  \end{proposition}

A key property of RDP is that it can be converted to standard $(\epsilon,\delta)$-DP using the Proposition~3 of~\cite{mironov2017renyi}, as follows:
\begin{proposition}
[From RDP to $(\epsilon,\delta)$-DP~\cite{mironov2017renyi}]
\label{prop:rdptodp}
	If $\mathcal{A}$ is an $(\alpha,\epsilon)$-RDP algorithm, then it also satisfies $(\epsilon+\nicefrac{\log (1/\delta)}{\alpha-1},\delta)$-DP for any $\delta\in(0,1)$.
\end{proposition}

\section{Deferred Theoretical Arguments}\label{sec:proof}

\subsection{Proof of \autoref{thm:privacy}}\label{sec:proof:privacy}

\begin{proof}
	In \autoref{alg:training}, the graph's adjacency is only used when the NAP mechanism is invoked during the forward propagation of submodels $\mathcal{M}_1$ to $\mathcal{M}_K$. According to Lemma 1 of~\cite{sajadmanesh2023gap}, the edge-level sensitivity of the NAP mechanism is 1, and thus based on Corollary~3 of~\cite{mironov2017renyi}, each individual query to the NAP mechanism is $(\alpha, \nicefrac{\alpha}{2\sigma^2})\text{-RDP}$. Due to \name{}'s caching system, the NAP mechanism is only invoked $K$ times during training (once for each submodel), and the rest of the training process does not query the graph edges. As a result, \autoref{alg:training} can be seen as an adaptive composition of $K$ NAP mechanisms, which based on Proposition~\ref{prop:rdpcompose}, is $(\alpha, \nicefrac{K\alpha}{2\sigma^2})\text{-RDP}$. According to Proposition~\ref{prop:rdptodp}, this is equivalent to edge-level $(\epsilon, \delta)$-DP with $\epsilon=\frac{K\alpha}{2\sigma^2} + \frac{\log(1/\delta)}{\alpha-1}$. Minimizing this expression over $\alpha>1$
	gives $\epsilon = \frac{K}{2\sigma^2} + \nicefrac{\sqrt{2K\log{(1/\delta)}}}{\sigma}$.
\end{proof}

\subsection{Proof of \autoref{thm:nodeprivacy}}\label{sec:proof:nodeprivacy}

\begin{proof}
	\autoref{alg:training} is composed of $K+1$ stages, where each stage $s\in\{1,\dots,K\}$ starts by computing and perturbing the aggregate embeddings (\autoref{eq:nap}), which is the only part where the graph adjacency information is involved. As this part is privatized by the NAP mechanism, the rest of the process in stage $s\ge1$ is just normal graph-agnostic training over tabular-like data, which is made private using DP-SGD. The exception is stage 0, which does not use the graph's adjacency  at all, and thus it is just privatized using DP-SGD. Therefore, \autoref{alg:training} can be seen as an adaptive composition of $K$ NAP mechanisms and $K+1$ DP-SGD algorithms. According to Lemma 3 of~\cite{sajadmanesh2023gap}, the NAP mechanism is node-level $(\alpha, \nicefrac{D\alpha}{2\sigma_{AP}^2})\text{-RDP}$. The DP-SGD algorithm itself is a composition of $T$ subsampled Gaussian mechanisms, which according to Theorem 11 of~\cite{pmlr-v97-zhu19c} and Proposition~\ref{prop:rdpcompose} is $(\alpha, \epsilon_\text{DPSGD})\text{-RDP}$, where:
  \begin{multline*}
    \epsilon_\text{DPSGD} \le \frac{T}{\alpha-1}\log\left\lbrace \left(1-\frac{B}{N}\right)^{\alpha-1}\left( \alpha\frac{B}{N} - \frac{B}{N} + 1 \right) \right. \\
     \left. + \binom{\alpha}{2}\left(\frac{B}{N}\right)^2\left(1-\frac{B}{N}\right)^{\alpha-2}e^{\frac{C^2}{\sigma_{GP}^2}} \right. \\
     \left. +\sum_{l=3}^\alpha\binom{\alpha}{l}\left(1-\frac{B}{N}\right)^{\alpha-l}\left(\frac{B}{N}\right)^l e^{(l-1)(\frac{C^2 l}{2\sigma_{GP}^2})} \right\rbrace.
    % & + \frac{DK\alpha}{2\sigma_{AP}^2} + \frac{\log(1/\delta)}{\alpha-1},
  \end{multline*}
  Overall, according to Proposition~\ref{prop:rdpcompose}, the composition of $K$ NAP mechanisms and $K+1$ DP-SGD algorithms is $(\alpha, \epsilon_{\text{total}})\text{-RDP}$, where:
  \begin{multline*}
    \epsilon_\text{total} \le \frac{(K+1)T}{\alpha-1}\log\left\lbrace \left(1-\frac{B}{N}\right)^{\alpha-1}\left( \alpha\frac{B}{N} - \frac{B}{N} + 1 \right) \right. \\
    \left. + \binom{\alpha}{2}\left(\frac{B}{N}\right)^2\left(1-\frac{B}{N}\right)^{\alpha-2}e^{\frac{C^2}{\sigma_{GP}^2}} \right. \\
    \left. +\sum_{l=3}^\alpha\binom{\alpha}{l}\left(1-\frac{B}{N}\right)^{\alpha-l}\left(\frac{B}{N}\right)^l e^{(l-1)(\frac{C^2 l}{2\sigma_{GP}^2})} \right\rbrace
    + \frac{DK\alpha}{2\sigma_{AP}^2}.
  \end{multline*}
  The proof is completed by applying Proposition~\ref{prop:rdptodp} to the above expression and minimizing the upper bound over $\alpha>1$:
  \begin{multline*}
    \epsilon \le \min_{\alpha>1} \frac{(K+1)T}{\alpha-1}\log\left\lbrace \left(1-\frac{B}{N}\right)^{\alpha-1}\left( \alpha\frac{B}{N} - \frac{B}{N} + 1 \right) \right. \\
    \left. + \binom{\alpha}{2}\left(\frac{B}{N}\right)^2\left(1-\frac{B}{N}\right)^{\alpha-2}e^{\frac{C^2}{\sigma_{GP}^2}} \right. \\
    \left. +\sum_{l=3}^\alpha\binom{\alpha}{l}\left(1-\frac{B}{N}\right)^{\alpha-l}\left(\frac{B}{N}\right)^l e^{(l-1)(\frac{C^2 l}{2\sigma_{GP}^2})} \right\rbrace \\
    + \frac{DK\alpha}{2\sigma_{AP}^2} + \frac{\log(1/\delta)}{\alpha-1},
  \end{multline*}
\end{proof}

\end{document}